\documentclass[11pt,a4paper]{article}

\usepackage{graphicx}             
\usepackage[accepted]{easrp2026}  

\usepackage{enumitem}
\usepackage{float}
\usepackage{subcaption}

\usepackage{cleveref}
\usepackage{mdframed}
\global\mdfdefinestyle{mydraft}{linecolor=red, linewidth=3pt, leftline=true, rightline=true, topline=false, bottomline=false}
\usepackage{soul}
\setulcolor{red}
\usepackage{placeins}

\newcommand{\ie}{i.e.\ }

\usepackage{listings}

\newcommand\blfootnote[1]{%
  \begingroup
  \renewcommand\thefootnote{}\footnotetext{#1}%
  \endgroup
}

\easrpauthor{
\vspace{1em}
    \textbf{\small Davide Venuto}\textsuperscript{1},
   \textbf{\small Leonardo Venuto}\textsuperscript{2}
}
\easrpaffiliation{
    \textsuperscript{1} {\small Technical University of Denmark (DTU), Lyngby, Denmark, \textup{davide.venuto011@gmail.com}}\\
    \textsuperscript{2} {\small Independent Researcher, Heidelberg, Germany, \textup{leonardo.venuto.ml@gmail.com}}
}
\easrptitle{
Implicit Reasoning in Transformers: Bridging the Compositionality Gap in GPT-2-like Models
}

\begin{document}

\maketitle

\vspace{-0.2cm}

\begin{abstract}
Compositional generalization remains a persistent limitation of transformer language models: even when a model learns to combine known facts correctly in-distribution (ID), the same compositional mechanism may fail when applied to out-of-distribution (OOD) knowledge. We investigate this compositionality gap in a controlled two-hop reasoning setting using a GPT-2-like transformer model trained from scratch. Building on prior work identifying a circuit associated with compositional generalization, we test whether the failure of OOD transfer is related to how atomic knowledge is structurally exposed during training. 
We modify the training distribution by presenting both ID and OOD atomic facts in a format positionally aligned with the second-hop retrieval step of the learned compositional circuit, without modifying the model architecture or adding OOD compositional supervision. Under this intervention, the model successfully transfers the learned ID compositional behaviour to held-out OOD atomic facts,
closing the observed compositionality gap in the controlled setting.
Our results provide evidence that the observed compositionality gap is not a fundamental architecture limitation and that it can be overcome through mechanistically informed training distribution design.

\end{abstract}

\blfootnote{\hspace{-0.5em}\textit{Code available at:} \url{https://github.com/Davide011/implicit-reasoning-bridging-compositionality-gap}}\vspace{-0.2cm}

\section{Introduction}\enlargethispage{0.5\baselineskip}
Large language models (LLMs) can store vast amounts of knowledge,
but recalling individual facts is rarely enough, many questions require combining several pieces of information through \textit{multi-hop reasoning} \citep{Wei2023,Press2023,Wang2024}. 
When this composition happens inside a single forward pass, without verbalizing intermediate steps, it is called \emph{implicit} reasoning. 
This step is all but trivial, it has been shown that models that can reliably answers two atomic queries will often fail on their composition \citep{Press2023,Yang2025}, a failure \citet{Press2023} called the \textit{compositionality gap}.\\

Explicit decomposition methods such as Chain-of-Thought can narrow the compositionality gap by verbalizing intermediate steps \citep{Wei2023,Press2023}, an inference-time solution that acts on top of the core capabilities the model acquires during (pre-)training \citep{Zhou2023,Lin2023}.\\

Controlled studies show that grokked transformers (\ie trained far beyond overfitting) can learn implicit two-hop composition tasks. For example, given the atomic facts “Barack’s wife is Michelle” and “Michelle was born in 1964,” the model can infer that “Barack’s wife was born in 1964.” These models generalize to unseen In-Distribution (ID) cases (atomic facts that appear in compositional training examples), yet fail to systematically transfer the same compositional behaviour to Out-of-Distribution (OOD) cases (atomic facts seen only in isolation, never in compositional training examples) 
\citep{Wang2024,Karmim2026,Yang2025,Biran2024}. \citet{Ye2025} pushed this further by showing that generalization extends to queries whose first-hop is OOD, but never to those whose second-hop is. So the natural question is: what limits the transfer of a learned compositional mechanism to OOD atomic facts, and under what training-data conditions can such generalization emerge?\\

We address this question by modifying the training distribution while keeping the task and model architecture fixed, without introducing any additional compositional supervision, working with the same rigorous synthetic setup from \citet{Wang2024}. Specifically, we introduce a position-aligned atomic-fact format that exposes atomic facts in the configuration associated with second-hop processing, and use controlled ablations to identify the conditions under which the learned compositional behaviour transfers to OOD knowledge, enabling OOD compositional generalization.\\

The main contributions of this work are as follows:
\begin{itemize}
\item We show that exposing atomic facts at the position corresponding to the second hop is insufficient  on its own  for OOD compositional rule transfer, demonstrating that positional alignment alone does not close the compositionality gap.

\item We identify an additional training-distribution condition associated with the successful OOD compositional transfer: a shared position-aligned representation across ID and OOD atomic facts that implicitly connects the learned ID compositional rule to OOD facts.
\end{itemize}

\section{Related work}

\paragraph{The compositionality gap.}
Prior research suggests that increasing model scale alone does not solve the compositionality gap \citep{Press2023} and evidence points to the second hop as the bottleneck: \citet{Yang2025} find that indications of latent first-hop resolution strengthen with model size while second-hop performance remains largely unaffected.
By training transformers from scratch on a synthetic knowledge graph, \citet{Wang2024} recover the circuit underlying composition: lower layers resolve the first hop and store the found  intermediate entity connecting the two atomic facts (the bridge entity), while upper layers use this entity to retrieve the second-hop answer.
\citet{Karmim2026} find that compositional generalization is achieved only for entities exposed to compositional examples in training (ID). \citet{Ye2025} additionally find that compositional generalization extends to queries whose \emph{first-hop} is unseen in compositional training, however second-hop generalization requires prior exposure of the corresponding fact in the second-hop position. 
The same asymmetry appears in pretrained models: \citet{Biran2024} show that the bridge entity is resolved in early layers while the second hop is only resolved later, and that patching a later-layer representation back into an earlier layer recovers the correct answer in up to $66\%$ of previously failed queries.
Taken together, these findings suggest that models’ ability to combine learned knowledge depends on how that knowledge is presented during training.

\paragraph{Attempts to Close the Gap.}
One  possible approach to addressing the compositionality gap is architectural. \citet{Wang2024} show that a parameter-sharing transformer, similar to the Universal Transformer \citep{Dehghani2019},  can unlock OOD compositional generalization, motivating recurrence as a mechanism for sharing knowledge across layers. More recent work extends this direction further: recurrent-depth transformers achieve systematic OOD generalization and can extrapolate to greater reasoning depths by increasing inference-time recurrence \citep{Kohli2026}, while \citet{Fu2026} identify an additional representational bottleneck in looped transformers and introduce DiscoLoop to improve multi-hop generalization. These approaches modify the model's computational structure to support iterative reuse of information across depth. In contrast, we ask whether OOD compositional transfer can emerge in standard non-recurrent transformers through changes to the training distribution alone.

Closest to our setting, \citet{Lin2026} leave the architecture intact and use a simplified two-hop training dataset in which all possible second-hop facts are always observed in that positional role during training. They introduce a zero-hop identity-bridge objective that maps each bridge-entity to itself, encouraging alignment between its representation as the output of the first hop and as the input to the second, thereby enabling OOD two-hop predictions without compositional supervision. The authors attribute this effect to implicit regularization favouring a direct subject-to-answer shortcut rather than genuine two-step inference.
In contrast, our intervention is guided by the previously found  compositional generalization circuit \citep{Wang2024} and aims to transfer an existing compositional mechanism to OOD atomic facts, rather than inducing a direct shortcut mapping.

\section{Methods}
Motivated by the previously hypothesized mismatch between the two-hop generalization circuit and OOD atomic-fact exposure, we intervene on the training distribution without modifying the model architecture or adding compositional supervision. Controlled ablations are then used to determine which components of the intervention are associated with successful OOD compositional generalization.

\subsection{Task and Data} 
\label{sec:task_data}
We build on the controlled two-hop composition task introduced by \citet{Wang2024}, where atomic facts are represented as directed relations $(h,r_1,b)$ and $(b,r_2,t)$, where $h$, $b$, and $t$ are respectively the head, bridge, and tail entities, and $r_i, i \in |\mathcal{R}|$ are the relations. Their composition requires predicting the tail entity $t$ from the query $(h,r_1,r_2)$: 
\begin{equation}
\forall h,b,t \in \mathcal{E},\ \forall r_1,r_2 \in \mathcal{R},\quad \quad \quad  \quad
(h,r_1,b)\land(b,r_2,t)\xrightarrow{\quad \text{composition rule} \quad}(h,r_1,r_2,t)
\end{equation}
For example, given the atomic facts “Barack's wife is Michelle” $(h,r_1,b)$ and “Michelle was born in 1964” $(b,r_2,t)$, the model must infer that “Barack's wife was born in 1964”. Here, Michelle is the bridge entity (b) connecting the two facts, and the model must predict the tail entity (t=1964) without receiving the bridge entity as part of the query.
Atomic facts are partitioned into in-distribution (ID; $95\%$) and out-of-distribution (OOD; $5\%$) subsets. Training contains all atomic facts and two-hop examples composed exclusively of ID facts, while OOD compositions are held out for OOD compositional evaluation. The model therefore observes the constituent OOD facts during training, but never their use within a compositional example. Our main experiments use the smaller and denser version of the synthetic graph, with $|\mathcal{E}|=200$ entities, $|\mathcal{R}|=20$ relation types, out-degree $20$ (\ie number of edges leaving each node), and inferred-to-atomic ratio $\phi=12.6$. Experiments on the original full-scale setting are reported in appendix (\cref{appendix:full_data_results}).

\subsection{Training-Distribution Intervention: Position-Aligned Atomic Exposure}  
\label{sec:new_data_format}
The intervention is motivated by the generalization circuit identified in prior work \citep{Wang2024,Biran2024}, where the first relation is resolved in the lower layers and the resulting bridge entity $b$ is subsequently combined with $r_2$ in the upper layers, and the observation that second-hop generalization requires the composition to appear during training \citep{Ye2025,Karmim2026}.
Under the original training distribution, OOD atomic facts are learned in isolation but are never presented in the input configuration corresponding to this second stage of composition. We  therefore introduce a \textit{position-aligned} data intervention, implemented by additionally presenting all the atomic facts in a \textit{second-hop-compatible} format,  designed to match the positional configuration associated with the second hop of the learned compositional mechanism, exposing both ID and OOD facts in the input position occupied by the second atomic relation during two-hop composition. Concretely, an atomic fact \cref{eq:atomic} is additionally presented as in \cref{eq:atomic_pos_aligned}:\\

\begin{minipage}{.4\linewidth}
    \begin{equation}
        \label{eq:atomic}
        (b,r_2,t) 
    \end{equation} 
\end{minipage}
\begin{minipage}{.4\linewidth}
    \begin{equation}
        \label{eq:atomic_pos_aligned}
        (\langle m\rangle,b,r_2,t) 
    \end{equation} 
\end{minipage}
\\

\noindent
where $\langle m\rangle$ is a non-predictive special token\footnote{  In the released implementation this token is denoted as \texttt{<mask>}; here we use $\langle m\rangle$ to avoid confusion with conventional masking tokens, such as those used in masked language models. The token serves as a positional placeholder and is excluded from the training loss.} 
used to align the atomic fact with the relevant input position. Crucially, these examples contain only the original atomic relation and introduce no additional two-hop compositional training examples. The intervention therefore changes how existing atomic knowledge is exposed during training rather than providing additional compositional supervision.

\subsection{Ablation Design}
\label{sec:ablation_design}
To isolate the training-distribution conditions required for OOD compositional transfer, we evaluate three controlled ablations of the full position-aligned intervention. Each ablation is applied independently, with all other components of the training distribution held fixed.\\

\begin{enumerate}[label=(\alph*), leftmargin=*, nosep]
    \item \textbf{Removing second-hop-compatible ID examples.}
    We remove all ID atomic examples presented in the new second-hop-compatible format while retaining their OOD counterparts, testing whether the shared representation is required for the compositional rule transfer.

    \item \textbf{Partial second-hop-compatible OOD exposure.}
    We retain all ID atomic examples in the new second-hop-compatible format while restricting this format to a subset of OOD atomic facts, testing the effect of partial OOD exposure.

      \item \textbf{Removing two-hop compositional supervision.} 
      We remove all two-hop compositional training examples while retaining position-aligned atomic facts, testing whether the modified atomic representation alone is sufficient to support compositional behaviour.
\end{enumerate}

\subsection{Model and Training Setup}
Following the setup presented by \citet{Wang2024}, a standard decoder-only transformer as in GPT-2 \citep{Radford2019language} was trained from scratch to avoid confounding effects from pretraining. The model consists of 8 layers, a hidden dimension of 768, and 12 attention heads, optimized using AdamW \citep{Loshchilov2019} with a learning rate of \(10^{-4}\), batch size 512, weight decay 0.03, and 2,000 linear warm-up steps. A dataset-specific vocabulary for the synthetic symbols was constructed using \textit{AutoTokenizer}.
Across all experiments, model performance is evaluated using exact-match accuracy on both in-distribution (ID) and out-of-distribution (OOD) two-hop compositions.

\section{Results}
We first evaluate whether the full position-aligned intervention recovers OOD compositional generalization, and then examine its components through controlled ablations.

\begin{figure}[h]  
    \centering
    \includegraphics[scale=0.5]
    {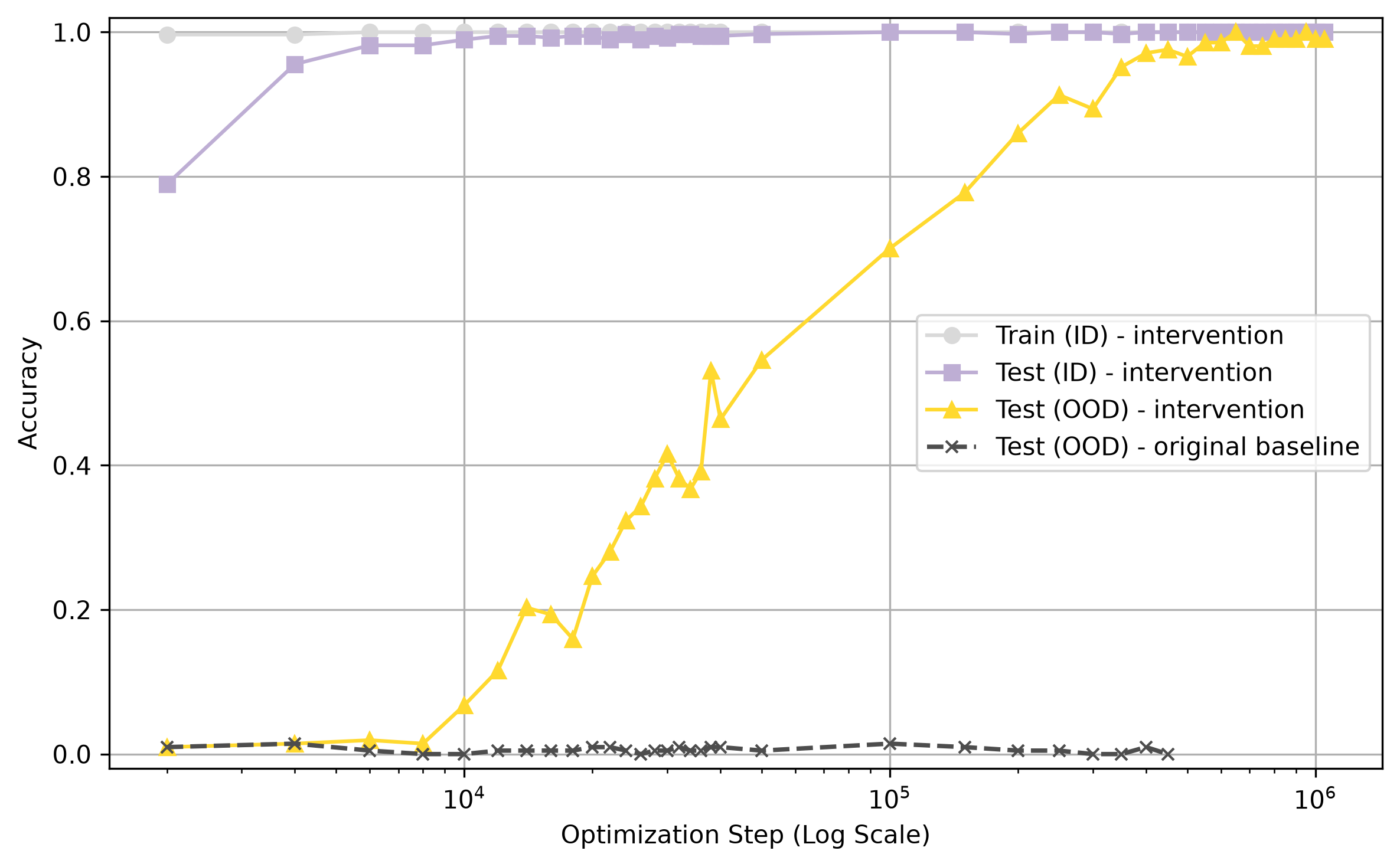}

    \caption{\textbf{OOD Compositional Generalization under the Full Position-Aligned Intervention.} ID and OOD test accuracy on the two-hop compositional task where both ID and OOD atomic facts are additionally presented in the second-hop-compatible format during training. OOD accuracy rises with delayed generalization in the grokking training regime, eventually reaching complete compositional generalization and closing the compositionality gap in the controlled experimental setting. The dashed gray curve shows OOD test accuracy under the original training distribution, without position-aligned atomic exposure, which remains near zero.}
    \label{fig:mask_B_generalization}
\end{figure}

\subsection{Recovering OOD Compositional Generalization}
The full position-aligned intervention presents inside the training dataset both ID and OOD atomic facts in the second-hop-compatible format introduced in Section~\ref{sec:new_data_format}. As can be seen in \cref{fig:mask_B_generalization}, we find that the model succeeds in generalizing the compositional rule learned from ID examples to OOD compositional cases, where the constituent atomic facts are never observed in compositional examples during training, closing the compositionality gap in the controlled setting. 
This result shows that OOD compositional generalization can be achieved with the modified training distribution.

\subsection{Ablation Analysis}
\label{sec:ablations}
The following ablations isolate which components of the position-aligned intervention are required for OOD compositional transfer. Each ablation modifies one component of the full intervention while leaving the remaining training setup unchanged.\\

\begin{figure}[h]
    \centering
    \setlength{\tabcolsep}{1pt}
    \renewcommand{\arraystretch}{0}
    \begin{subfigure}[b]{0.4\textwidth}
        \centering
        \includegraphics[width=\textwidth]{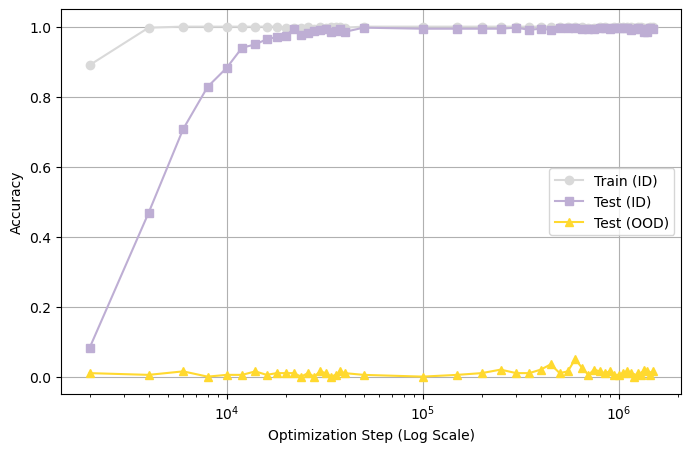}
        \caption{\textbf{No ID Atomic Facts in the Second-Hop-Compatible Format}}
        \label{fig:ood_only}
    \end{subfigure}
    \hspace{-0.05cm} 
    \begin{subfigure}[b]{0.4\textwidth}
        \centering
        \includegraphics[width=\textwidth]
        {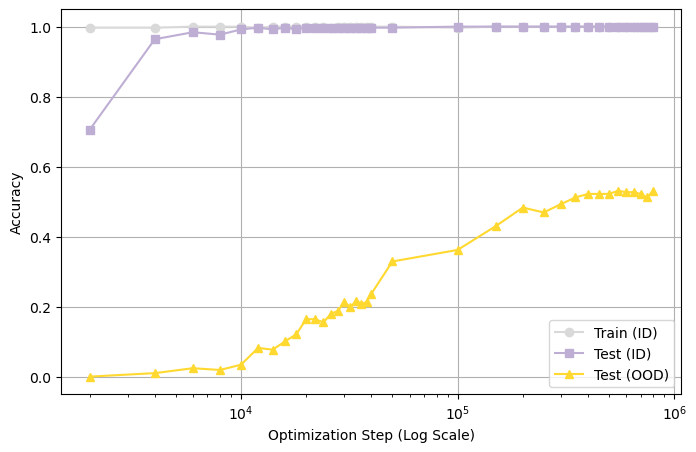} 
        \caption{\textbf{Partial OOD Atomic Facts in the Second-Hop-Compatible Format}}
        \label{fig:partial_ood}
    \end{subfigure}
    \hspace{-0.05cm} 
    \begin{subfigure}[b]{0.4\textwidth}
        \centering
        \includegraphics[width=\textwidth]
        {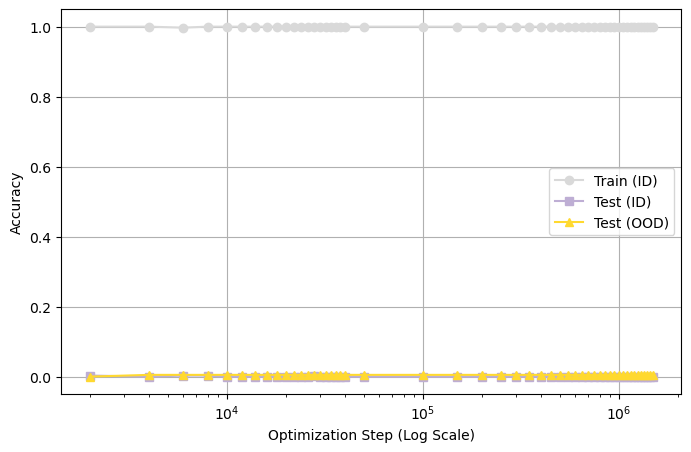}
        \caption{\textbf{No 2-Hop Composition Examples}}
        \label{fig:no_2_hop}
    \end{subfigure}

    \caption{\textbf{Generalization Performance Across Ablation Experiments.}
    Comparison of ID and OOD test accuracy on the two-hop compositional task across three controlled ablations.
    (a) Removing ID atomic facts from the second-hop-compatible format prevents OOD compositional rule transfer.
    (b) Restricting the second-hop-compatible format to only a subset of OOD atomic facts results in partial OOD compositional generalization.
    (c) Removing two-hop compositional training examples prevents the model from learning the compositional task for both ID and OOD atomic facts.}
    \label{fig:ablation_comparison}
\end{figure}

\begin{enumerate}[label=(\alph*), leftmargin=*, nosep]
    \item \textbf{Second-Hop OOD Exposure Is Insufficient for Compositional Transfer}

We test removing all ID atomic facts from the second-hop-compatible format while retaining the corresponding OOD examples. This preserves position-aligned exposure of OOD knowledge but removes the shared format between the ID atomic knowledge supporting compositional training and the OOD atomic knowledge on which the compositional transfer is evaluated.
In \cref{fig:ood_only} we can see that in this case OOD compositional generalization is not achieved, therefore exposing OOD atomic facts in the second-hop-compatible format is not sufficient by itself to enable compositional transfer. 
Successful OOD generalization additionally requires ID atomic facts to be presented in the same format.

  \item \textbf{Partial Second-Hop OOD Exposure}

We retain all ID atomic facts in the second-hop-compatible format while restricting this format to only a subset of OOD atomic facts. As shown in Figure~\ref{fig:partial_ood}, this intervention produces only partial OOD compositional generalization, in contrast to the complete recovery observed when all OOD atomic facts receive position-aligned exposure.
This result indicates that the extent of second-hop-compatible OOD exposure directly constrains the extent of compositional transfer. In the full-scale setting, we observe that partial exposure is insufficient to recover OOD generalization (see appendix \cref{appendix:ablation_full_model}, \cref{fig:mask_b_not_all_ood}). 

\pagebreak
\item \textbf{Removing Compositional Supervision}

Lastly we test retaining the second-hop-compatible format for both ID and OOD facts while removing all two-hop compositional examples from the training dataset. In this case we observe that the model fails to generalize the compositional task, not just for the OOD set, but for the ID test set as well (\cref{fig:no_2_hop}).
Additionally we also test the identity-bridge supervision proposed by \citet{Lin2026}, they used the same no-compositional supervision condition as in this ablation step, but we do not achieve compositional generalization in our higher-complexity setting; full results are reported in appendix (\cref{appendix:experiment_identy_bridge}).\\

\end{enumerate}

\noindent
Taken together, the ablations show that atomic facts second-hop-compatible exposure alone is insufficient for OOD compositional generalization. Complete recovery is observed when the position-aligned format is shared by both ID and OOD atomic facts while compositional supervision on ID examples is retained.

\section{Discussion}
Our ablation analysis suggests that the compositionality gap cannot be attributed solely to the lack of second-hop-compatible exposure. Rather, successful OOD generalization appears to depend on establishing a connection between OOD atomic facts and ID ones, for which compositional context is learned.
We hypothesize that the shared position-aligned format provides such a connection at the second-hop stage: ID facts link the compositional training to the new format, which is then shared with OOD facts. This overlap may bias optimization toward representations in which ID and OOD knowledge can be processed in the same way by the learned second-hop computation, allowing compositional context learned on ID examples to extend to OOD cases.
This interpretation is consistent with the mechanism observed by \citet{Ye2025}, where OOD first-hop representations become aligned with second-hop ID representations, enabling this partial cross-distribution generalization.\\

Another perspective is provided by the identity bridge method introduced by \citet{Lin2026}, which shows that bridge entity supervision can bias optimization towards shortcut-like subject-to-answer mappings, producing OOD two-hop predictions even without explicit compositional training for sufficiently simple datasets. This suggests a common role for training-distribution-structure in steering optimization toward representations that support cross-example generalization. The resulting behaviour, however, differs in the two settings. \citet{Lin2026} analyse when such structure can induce a compositional-looking mapping from atomic supervision alone, whereas our intervention 
aims at connecting ID and OOD facts via the second-hop-aligned format, enabling ID compositional rule to be generalized to OOD facts. Consistent with their reported results degrading as task complexity increases, we find that identity bridge supervision does not induce any compositional generalization in our higher-complexity setting without two-hop supervision (see appendix \cref{appendix:experiment_identy_bridge}).
While differences between the experimental settings prevent a direct comparison, this control suggests that our successful intervention operates in a different way: rather than inducing a new shortcut from atomic supervision, it manages to transfer the compositional rule learned from ID examples to OOD facts.\\

More broadly, our results suggest that OOD rule transfer can depend strongly on how knowledge is organized and exposed during training, rather than on architectural capacity alone. This is particularly relevant in settings where atomic knowledge is available but exhaustive supervision over its possible compositions is infeasible. Our findings suggest that training-distribution design may play a role in connecting learned compositional mechanisms to knowledge not encountered in compositional training.

\subsection{Limitations \& Future Directions}
The present study is restricted to two-hop compositional reasoning within a controlled GPT-2-like setting and does not establish whether the observed generalization behaviour extends to longer reasoning chains. Moreover, the experiments rely on synthetic, algorithmically generated data, which enables precise control over the training distribution but limits the direct applicability of the findings to natural-language and real-world reasoning settings.\newline

We expect the broader principle of facilitating connections between learned compositional mechanisms and OOD knowledge through training-distribution design to remain relevant beyond this setting. However, in more complex and less structured tasks, establishing such connections may require more targeted interventions to guide optimization. Furthermore, in larger models and less structured settings, compositional mechanisms may be more distributed and their computational stages less clearly identifiable, making the relevant mechanisms and training conditions harder to identify. While the specific position-aligned format introduced here may not transfer directly, mechanistic analysis could help identify compositional generalization bottlenecks and guide appropriate interventions in more complex settings. Further work is therefore required to determine whether the identified training conditions generalize across greater compositional depth, model scales, and more complex data distributions.

\section{Conclusion}
While explicit reasoning approaches such as Chain-of-Thought can help mitigate the compositionality gap \citep{Wei2023,Press2023}, they do not eliminate the challenges of reasoning over large search spaces. \citet{Wang2024} demonstrate that, in a complex comparison task, transformers leveraging knowledge encoded in their parameters can succeed where Chain-of-Thought approaches struggle. Therefore understanding how such implicit reasoning capabilities emerge and generalize remains important, alongside approaches that explicitly decompose reasoning into intermediate steps.\\

In this work, we show that the compositionality gap can be closed in a controlled two-hop reasoning setting without architectural modifications or additional compositional supervision, solely by changing how atomic facts are exposed during training.
Crucially, we find that second-hop-compatible exposure of OOD facts is not sufficient on its own to successfully transfer the compositional rule. OOD compositional generalization emerges when ID and OOD atomic facts share the second-hop position-aligned representation, connecting OOD facts to the ID examples for which the compositional rule is learned. Under this training condition, a GPT-2-like transformer generalizes its learned ID compositional behaviour to OOD atomic facts. These findings suggest that failures of OOD compositional transfer depend critically on the training distribution, rather than solely on architectural capacity.

\enlargethispage{\baselineskip}
\section*{Acknowledgments}
The authors thank Professors Ole Winther and  Thomas Bolander for their guidance and insightful discussions during the  thesis research on which this paper is based, and for their continued advice. We thank the EuroTech Federation for supporting further development of this research  through the European Summer Research Program and acknowledge DTU Compute for access to the Titans HPC cluster and Amazon Web Services (AWS) for additional computational resources.

\pagebreak

\bibliographystyle{plainnat}
\bibliography{references}
\clearpage
\appendix
\section{Appendix} 

\subsection{Dataset Creation Details}
\label{appendix:dataset_creation}
The dataset for 2-hop reasoning is manually created using a graph construction approach. This is done for the sake of algorithmic simplicity, aligning with the experiment's goal of evaluating the model's ability to perform 2-hop reasoning on learned tokens. To keep the model as simple and computationally efficient as possible, both for training and for observing Grokking effects, the experiment relies on this highly simplified, constructed reasoning dataset presented in the analysed reference paper \citet{Wang2024}.

\subsubsection{Dataset Creation}
\label{appendix:full_dataset}

The data creation process follows a well-defined algorithmic structure, enabling controlled dataset generation. This approach allows for the careful artificial construction of in-distribution (ID) and out-of-distribution (OOD) atomic facts.
This is done by first creating the triplets that compose the atomic facts and splitting them into ID (95$\%$) and OOD atomic facts (5$\%$), following the format \textit{(entity, relation, entity)}. Then, the\textit{ two-hop inferred facts} are generated by matching atomic facts where \textit{the tail of one aligns with the head of another}, successfully constructing all possible 2-hop relations. These are then divided into ID cases (where both atomic facts are ID) for training purposes and OOD cases (where both atomic facts are OOD), which are reserved for evaluation and testing.  
Finally, the training and test datasets are created. 

\subsubsection{Position-Aligned Data Intervention} \label{appendix:new_data_format}
Position-aligned data intervention as described in \cref{sec:new_data_format}. Atomic facts are presented in a second-hop-compatible format by extending the original dataset construction code seen above, iterating over both ID and OOD facts, and restructuring them into the new format as described in the \cref{eq:mask_b_appendix}.

\begin{equation} \label{eq:mask_b_appendix}
\begin{aligned}
\underbrace{(<m>)}_{\substack{1^{\circ} position \\ \text{Special Token}}} \land 
\underbrace{(b, r_2, t)}_{\substack{2^{\circ} position \\ \text{Atomic Fact}}} 
\implies 
\underbrace{( <m> ,\hspace{1mm} b,\hspace{1mm} r_2, \hspace{1mm}t)}_{\substack{Atomics\hspace{1mm}at\hspace{1mm} 2^{\circ} Hop  \\ \text{-new format-}}}
\end{aligned}
\end{equation}

\subsubsection{Small vs Full Dataset} \label{working_data_small_appendix}

\begin{itemize}
    \item \textbf{Small-Data Model:} Refers explicitly to the model trained on the \textit{small working dataset} presented in the main section of the paper \cref{sec:task_data}.
    \item \textbf{Full Data Model:} Refers to the model trained on the full dataset described below in this section.
\end{itemize}

The main experiments presented in this paper use the small dataset described in \cref{sec:task_data}, with (200) entities and (20) relation types. The original full-scale setting introduced by \citet{Wang2024} follows the same data-generation procedure but increases the graph to (2{,}000) entities and (200) relation types, while keeping the out-degree fixed at (20).
This difference makes this paper main experimental setting substantially denser in terms of relation coverage. Each entity is assigned (20) distinct outgoing relations; therefore, in the smaller setting, each entity is associated with all (20) available relation types, whereas in the full-scale setting only (20) of the (200) possible relation types are assigned to each entity. The smaller setting consequently provides denser relational coverage per entity and empirically allows faster learning, making it suitable for the extensive intervention and ablation experiments reported in the main text. Results on the original full-scale setting are reported in appendix  (\cref{appendix:full_data_results}) to assess whether the main findings persist under the larger and sparser configuration.

\subsection{Identity-Bridge Experiment}
\label{appendix:experiment_identy_bridge}

\begin{figure}[H] 
    \centering
    \includegraphics[scale=0.5]{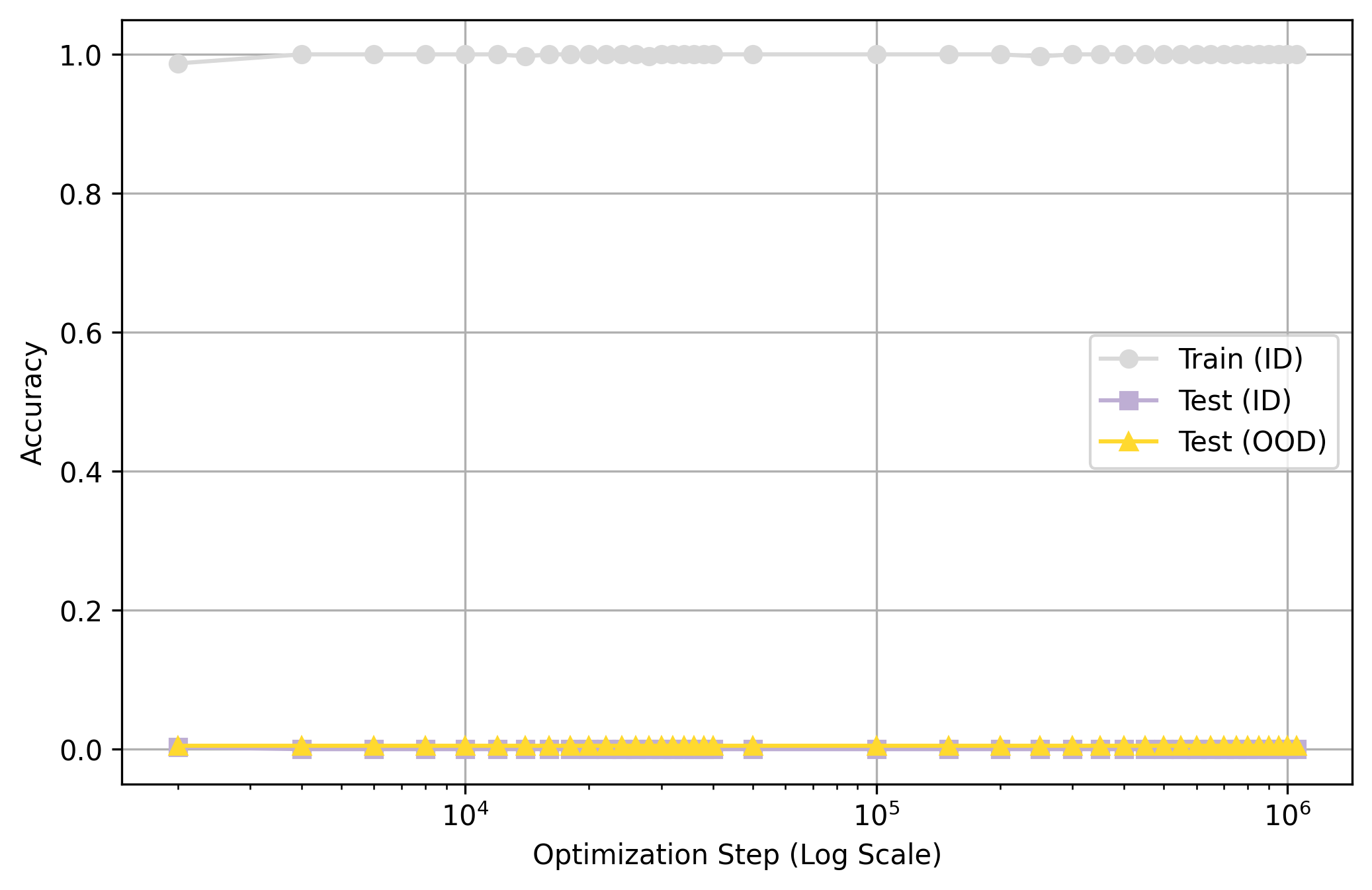}
    \caption{\textbf{Identity-Bridge Control without Compositional Supervision.}
Accuracy under identity-bridge supervision after removing all two-hop compositional training examples. While training accuracy remains near perfect, both ID and OOD compositional test accuracy remain at zero, showing that identity-bridge supervision alone does not induce compositional generalization in our setting.
    }
    \label{fig:identity_bridge_exp}
\end{figure}

As an additional control, we adapt the identity bridge intervention of
\citet{Lin2026} to our no-compositional-supervision setting. Because the
underlying dataset constructions differ, this experiment should be interpreted
as a conceptual adaptation rather than an exact reproduction.
\citet{Lin2026} define task complexity as the number of first-hop relation
types, \(C = |R_1|\), while fixing the second-hop relation set to
\(|R_2|=1\). Increasing \(C\) therefore increases branching only at the first
hop, with their reported experiments considering values up to \(C=9\).
Our setting contains \(20\) relation types that can occur at both hops.
Thus, when viewed along their first-hop complexity axis, our task has a larger
branching factor, while also introducing substantially greater second-hop
variability rather than fixing a single second-hop relation.
We introduce analogous zero-hop identity supervision while retaining the
atomic training data and removing all two-hop compositional examples. As shown
in Figure~\ref{fig:identity_bridge_exp}, the model fails to learn the compositional task. This outcome is consistent with the degradation reported by
\citet{Lin2026} as task complexity increases. In our setting, identity-bridge
supervision is therefore insufficient to induce compositional behaviour from
atomic supervision alone, further distinguishing shortcut induction from the
transfer of a compositional rule already learned from ID examples.

\subsection{Full-Data Model Results}
\label{appendix:full_data_results}

\subsubsection{Recovering OOD Compositional Generalization} 
\label{appendix:full_experiment}
\begin{figure}[H]
    \centering
    \setlength{\tabcolsep}{1pt}  
    \renewcommand{\arraystretch}{0}
    
    \begin{subfigure}[t]{0.48\textwidth}
        \centering
        \includegraphics[width=\textwidth]{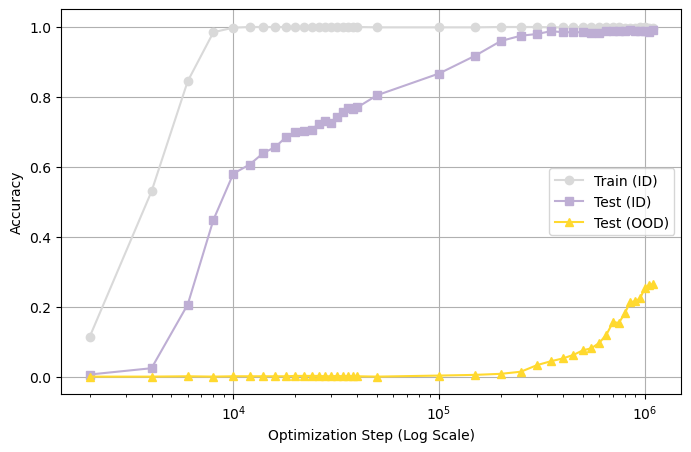}
        \caption{\textbf{Full-Dataset} }
        \label{fig:big_id_ood}
    \end{subfigure}
    \hspace{0.1mm} 
    \begin{subfigure}[t]{0.48\textwidth}
        \centering
        \includegraphics[width=\textwidth]{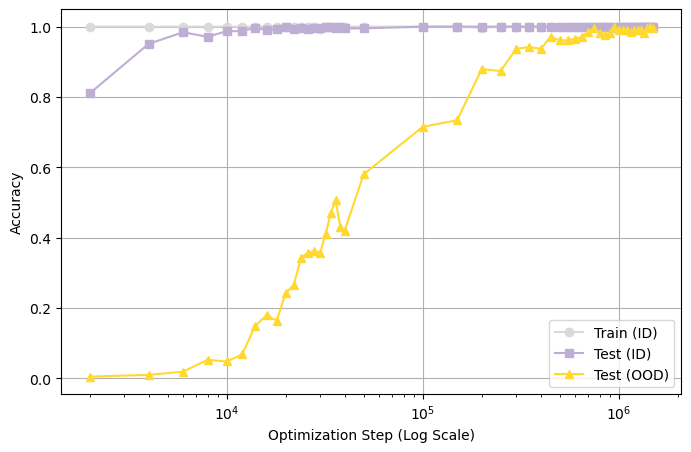}
        \caption{\textbf{Small-Dataset}}
        \label{fig:mask_generalization}
    \end{subfigure}

    \caption{\textbf{Impact of Dataset Size-Density on OOD Compositional Generalization}.\\
    (\subref{fig:big_id_ood}) When trained on the full but less dense dataset, the model demonstrates delayed convergence in both training and testing metrics. As a result, the generalization effect driven by grokking does not achieve full accuracy within the performed training steps. (\subref{fig:mask_generalization}) In contrast, training on a smaller but denser dataset leads to significantly improved generalization speed. The model reaches full accuracy on the OOD test set within the given optimization steps, highlighting the benefits of dataset density in accelerating generalization.}
    \label{fig:generalization_study}
\end{figure}

The primary distinction between the large and small datasets (both trained over an identical number of optimization steps) emerges clearly in the timing at which the model begins to exhibit grokking behaviour and generalize effectively to out-of-distribution (OOD) data, as can be seen in \cref{fig:generalization_study}. Specifically, the model trained on the larger ("full") dataset requires substantially more training time before demonstrating signs of generalization. Available evidence suggests that, with sufficient training, the model on the \textit{full dataset} would eventually achieve accuracy comparable to that of the model trained on the smaller dataset. In fact, nothing in the current evidence indicates a different long-term behaviour. However, \textit{this hypothesis remains untested in this research} due computational constraints.

\subsubsection{Ablation Analysis} 
\label{appendix:ablation_full_model}

\begin{figure}[H]
    \centering
    \setlength{\tabcolsep}{1pt}  
    \renewcommand{\arraystretch}{0}
    \begin{subfigure}[t]{0.48\textwidth}
        \centering
        \includegraphics[width=\textwidth]{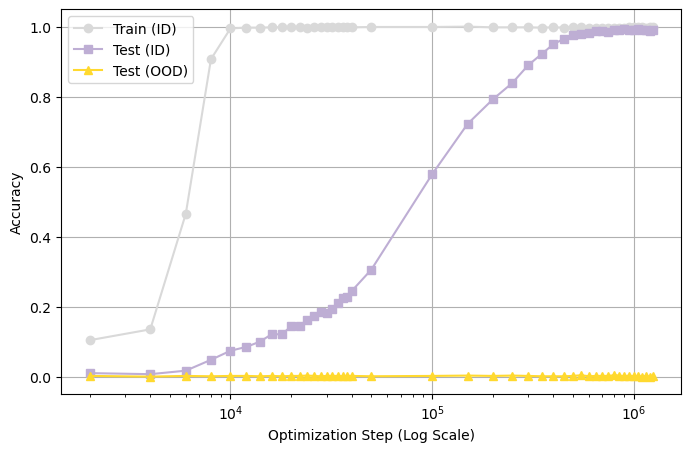}
        \caption{\textbf{No ID Atomics in the New   2° Position \newline Format }(full-data-model)}
        \label{fig:mask_big_no_id}
    \end{subfigure}
    \hspace{0.1mm} 
    \begin{subfigure}[t]{0.48\textwidth}
        \centering
        \includegraphics[width=\textwidth]{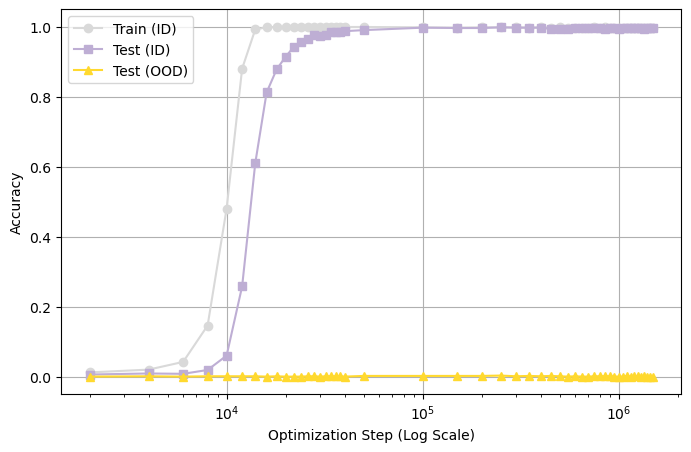}
        \caption{\textbf{Partial 2° Hop OOD Placement}\newline(full-data-model)}
        \label{fig:mask_b_not_all_ood}
    \end{subfigure}
    \caption{\textbf{Ablation Analysis Results on the Full-Model} for the experiments described in \cref{sec:ablation_design}.
    (\subref{fig:mask_big_no_id}): ablation experiment \textbf{(a)} where no in-distribution (ID) atomic facts are provided in the new 2°hop format. Under this condition, the Full Model similarly fails to generalize to OOD examples, resembling the behaviour previously observed in the Smaller Model under the same setting.
    (\subref{fig:mask_b_not_all_ood}): ablation experiment \textbf{(b)} where out-of-distribution (OOD) facts are partially placed in the second-hop position. In the Full Model experiment, this partial OOD placement appears to completely hinder OOD generalization. This contrasts with the behaviour observed in the Smaller Model, where the same setting resulted in only a partial reduction in final accuracy (\cref{fig:partial_ood}).
    }
    \label{fig:full_model_ablation_results}

\end{figure}

Results for the ablation analysis described in \cref{sec:ablation_design} performed on the \textbf{Full Data Model}.
Specifically: \textbf{(a)} examines the scenario where no \textit{ID atomic} facts are provided in the newly defined second-hop format, \textbf{(b)} corresponds to the partial placement of OOD facts in the second-hop position. Experiment \textbf{(c)} was not reproduced on the full dataset due to its clear failure already observed in the small dataset case.
\newline
\begin{enumerate}[label=(\alph*), leftmargin=*, nosep]
    \item In \cref{fig:mask_big_no_id} we can see that the \textit{Full data model} exhibits the same behaviour as observed in the smaller dataset (\cref{fig:ood_only}): OOD generalization is hindered due to the absence of the ID atomic facts in the new \textit{2°-hop format} during training.  This missing structure likely acts as a crucial \textbf{knowledge bridge}, and its exclusion disrupts the model's ability to transfer knowledge effectively across relational hops. This evidences its critical role in working as a \textit{knowledge bridge} between ID and OOD data from the model perspective.
    \item In \cref{fig:mask_b_not_all_ood} we can observe that the model trained on the full dataset fails to generalize to OOD facts in opposition of the small data model were it only hindered the reach of full accuracy given the set of missing atomic at the second hop (\cref{fig:partial_ood}). This limitation may be attributed to the lower density of the full dataset, where the reduction of OOD facts in the second-hop position appears critical for maintaining effective knowledge transfer.  However, it might also be possible that the generalization is delayed rather than entirely impeded. Further investigations are required to fully explain the observed results.
\end{enumerate}

\end{document}